\documentclass{ieeeaccess}
\usepackage{cite}
\usepackage{amsmath,amssymb,amsfonts}
\usepackage{algorithmic}
\usepackage{graphicx,color}
\usepackage{textcomp}
\usepackage{bbold}
\usepackage{algorithm}
\usepackage{array}
\usepackage[caption=false,font=normalsize,labelfont=sf,textfont=sf]{subfig}
\usepackage{textcomp}
\usepackage{stfloats}
\usepackage{url}
\usepackage{verbatim}
\usepackage{cite}
\usepackage{mathtools}
\usepackage{enumerate}
\usepackage{booktabs}
\usepackage{color}
\newcommand{\mat}[1]{\boldsymbol{#1}}
\newcommand{\real}{\mathbb{R}}
\newcommand{\norm}[1]{\left\lVert#1\right\rVert}

\def\BibTeX{{\rm B\kern-.05em{\sc i\kern-.025em b}\kern-.08em
    T\kern-.1667em\lower.7ex\hbox{E}\kern-.125emX}}
\begin{document}
\history{Date of publication xxxx 00, 0000, date of current version xxxx 00, 0000.}
\doi{10.1109/ACCESS.2023.0322000}

\title{Towards Resource-Efficient Federated Learning in Industrial IoT for Multivariate Time Series Analysis}
\author{\uppercase{Alexandros Gkillas}\authorrefmark{1}, \IEEEmembership{Student MEMBER, IEEE,},
\uppercase{Aris Lalos}\authorrefmark{1}
\IEEEmembership{Senior Member, IEEE}}

\address[1]{Industrial Systems Institute, Athena Research Center, Patras Science Park, Greece(e-mail: gkillas@isi.gr, lalos@isi.gr)}


\markboth
{Author \headeretal: Preparation of Papers for IEEE TRANSACTIONS and JOURNALS}
{Author \headeretal: Preparation of Papers for IEEE TRANSACTIONS and JOURNALS}

\corresp{CORRESPONDING AUTHOR: Alexandros Gkillas (e-mail: gkillas@isi.gr).}

\begin{abstract}

Anomaly and missing data constitute a thorny problem in industrial applications. In recent years, deep learning enabled anomaly detection has emerged as a critical direction, however the improved detection accuracy is achieved with the utilization of large neural networks, increasing their storage and computational cost. Moreover, the data collected in edge devices contain user privacy, introducing challenges that can be successfully addressed by the privacy-preserving distributed paradigm, known as federated learning (FL). This framework allows edge devices to train and exchange models increasing also the communication cost. Thus, to deal with the increased communication, processing and storage challenges of the FL based deep anomaly detection NN pruning is expected to have significant benefits towards reducing the processing, storage and communication complexity. With this focus, a novel compression-based optimization problem is proposed at the server-side of a FL paradigm that fusses the received local models broadcast and performs pruning generating a more compressed model. Experiments in the context of anomaly
detection and missing value imputation demonstrate
that the proposed FL scenario along with
the proposed compressed-based method
are able to achieve high compression rates (more than
$99.7\%$) with negligible performance losses (less than
$1.18\%$ ) as
compared to the centralized solutions.

\end{abstract}


\titlepgskip=-21pt

\maketitle
\section{Introduction}
\label{sec:intro}
In recent years, the Internet of Things (IoT) has revolutionized the industry discipline, paving the way for better efficiency, safety, and security in the manufacturing processes \cite{IIoT1}. Industry 4.0 emerged recently, as an efficient paradigm to handle the need of the inter-connectivity of Industrial IoT (IIoT), enabling the access to real time datasets derived from dispersed edge devices, which can sense the environment and process data  in an autonomous and decentralized manner \cite{IIoT2}. Nonetheless, the deployment of IoT devices in the Industrial domain introduces some crucial challenges.  To be more specific, the quality of the derived multidimensional data is often degraded by various factors e.g., faulty sensors and communication failures, thus introducing various types of anomalies (e.g. data instances that significantly deviate from the majority of data instances such as missing values and/or outliers) \cite{anomaly_review, anomaly, multidimensional, 9976093} and affecting heavily the performance  of various IIoT tasks, such as classification \cite{classification}, prediction \cite{forecasting}. 
To address the problem of anomaly detection and restoration 
numerous centralized solutions has been developed over the years. The considered problem has been studied under various scenarios and settings. Especially, utilizing the advances in deep learning, data-driven models including  RNN \cite{nature}, GAN \cite{GANa}, Transformers\cite{transformer} and CNN \cite{cnn_anomaly} have been
getting attention achieving state-of-the-art results in the considered problem.  Nonetheless, the above deep learning models require massive amounts of training data and significant computing and storage resources, thus rendering them unsuitable for IIoT edge devices. More importantly, in several cases the available data produced from  a single industrial site may be insufficient for learning accurate machine/deep learning models to address efficiently Anomaly detection, a.k.a. outlier detection or novelty detection \cite{AnomalyInd,AnomalyIoT}. An intuitive solution to tackle the above issue is to gather data collected from different parties and/or design at the same time more compact NN models. However, due to privacy constraints \cite{IIoT2}, the industrial entities may be reluctant to expose their owned dataset, while still being interested in an AI model trained using their own and privately owned datasets of others. 

Towards this direction, Federated learning (FL), as a secure distributed framework is capable of addressing the above challenges \cite{McMahan2017CommunicationEfficientLO}, enabling the clients  to collaboratively learn a deep learning model without sharing their own private datasets \cite{icassp1, 10314010, 10167897}. The privacy-preserving and distributed nature of the FL promotes such solutions at the network edge, where each IIoT device can be considered as a client containing one or more sensors that measure different physical quantities, thus recording a part of the generated multivariate data instances. Both the FL principle, that is based on the frequent exchange of trained models between the edge devices and the server, but also the limited computational and storage resources in IIoT devices render the compression and acceleration of the models an imperative requirement.



Thus our work focuses on providing an efficient framework for multivariate time series data by  utilizing compressed NNs at the edge following the FL paradigm, allowing the training of the models from a small subset of data instances. It is important to note that existing literature has predominantly focused on less practical scenarios in which edge devices or clients possess sensors that measure identical physical quantities, leading to a shared feature space \cite{fl_a1, fl_a3, fl3, fl4, fl5, fl6, fl7}. However, these scenarios may not be feasible in real-world situations, as different clients may have distinct sensors measuring different features. To address this limitation, our study explores a more realistic federated learning scenario, wherein each client is equipped with a unique sensor that measures only a portion of the multidimensional (multivariate) time series data, resulting in univariate data owned by the clients. Through the use of FL, the clients collaboratively aim to capture the multivariate information by integrating their individual univariate data, effectively reconstructing the full spectrum of information from the diverse sensor inputs.

 Furthermore, inspired by the compression and acceleration techniques aiming to reduce the size of deep learning networks \cite{8253600, Ac1, Ac2, Ac3, 10290057} and  considering the limitations of the IIoT edge devices in terms of computational and power resources \cite{IIoT2, iot},  a novel
compression-based optimization problem is proposed at the
server-side in order to fuse the local models broadcast by the
edge devices, thus deriving a compressed global model with
reduced number of weights without simultaneously affecting its performance accuracy, providing compression rates (weight pruning) greater than ${99.7\%}$ while preserving at the same time the performance.

To sum up, the key contributions of this paper are the following:
\begin{itemize}
    \item A  realistic federated learning  
     scenario is proposed considering that the edge devices contain sensors that measure only a part of multidimensional time series data. In other words, from univariate data 
    \item A novel compressed based fusion rule is proposed at the server-side to combine the local models of the edge devices, providing compressed global models with high compression rates  and no performance accuracy losses.
    \item  Extensive evaluation studies in the context of  anomaly detection and missing value imputation demonstrate that the proposed federated learning scenario along with the proposed compressed-based optimization problem are able to achieve high compression rates (more than $99.7\%$) with negligible performance losses (less than $1.18\%$ and $5\%$) for the two considered problems as compared to the centralized solutions. 
\end{itemize}


\section{Related works and Preliminaries}

In this section, we provide an overview of the existing literature on Federated Learning (FL) in multivariate time series data.

\subsection{Related works} \label{related}

Several studies have explored the use of FL for multivariate time series data analysis, aiming to improve the efficiency and accuracy of data processing in distributed settings. These works have demonstrated the potential benefits of FL in various applications \cite{fl_a1, fl_a3, fl3, fl4, fl5, fl6, fl7}. In more detail study in \cite{fl_a1, fl7} focused on a federated learning scenario for machinery fault diagnosis using the FedAvg algorithm. Study in \cite{fl_a3} introduced a GAN-based imputation method under the FL framework to solve the missing value imputation problem for multivariate data. Similar to the previous studies, method \cite{fl3, fl4, fl5} used the FL method to detect anomalies in  multidimensional time series datasets. 
However,  these studies assume a shared feature space among the clients or edge devices, limiting the practicality of their approaches in real-world scenarios where different clients may possess distinct sensors measuring diverse features. Moreover, the aforementioned approaches utilize different versions of the fedAvg algorithm \cite{fedAvg}, neglecting the fact that training large-scale deep learning models via Federated Learning can be computationally  for edge devices with limited resources.

Motivated by these limitations, in this work, we develop an efficient approach for handling multivariate time series data in a federated learning context, focusing on the practicality and resource efficiency of compressed deep learning models in edge computing environments. Our approach consists of two primary strategies: initially, by concentrating on univariate time series data from diverse sensors, we inherently simplify the data and reduce the model input data size, making the deep learning models more efficient, especially in resource-constrained edge computing environments. Secondly, we incorporate a compression mechanism specifically designed to compress the deep learning models while retaining the crucial information necessary for accurate analysis of multivariate data. This compression is performed on the server side during the fusion of client models in the FL process.

The proposed approach offers a more practical solution for multivariate time series data analysis in federated learning contexts by considering diverse sensors and features in real-world scenarios. Although the focus is on univariate time series data, it effectively utilizes the multivariate information through the federated learning process. This enables the approach to achieve comparable results to those obtained using multidimensional data, while benefiting from the computation efficiency and resource advantages offered by the univariate data inputs. By incorporating a model compression mechanism and considering real-world constraints, the proposed method delivers improved resource efficiency, and adaptability in various applications and settings, making it more suitable for real-world applications than other existing methods in the field.

\subsection{Preliminaries - Federated Learning Flow}
The federated learning (FL) methodology constitute an iterative process employing several communication rounds between the involved entities, i.e., the centralized server and the edge devices (users or clients). In this distributed framework, the main goal of the FL is to enable the participated edge devices to collaboratively learn a global machine/deep learning model without sharing any information regarding their private local datasets. In more detail, each communication round of index $m$ consists of the following recursively steps \cite{eldar}:
\begin{enumerate}
  \setlength{\itemsep}{1pt}
  \setlength{\parskip}{0pt}
  \setlength{\parsep}{0pt}
    \item \textit{Local Training}: Each  device $i$ uses its private local dataset to optimize a local model $\vartheta_i^m$.
    \item \textit{Server Aggregation}: The participated edge devices convey their updated local models to the server. Subsequently, the server employs some fusion (aggregation) rule to derive a global model $\vartheta^m$. 
    \item \textit{Model Distributing}: The server broadcast the new global model $\vartheta^m$ to the  devices, and the next iteration (round) $m+1$ is performed.
\end{enumerate}

\begin{figure*}
\centering
\includegraphics[scale=0.4]{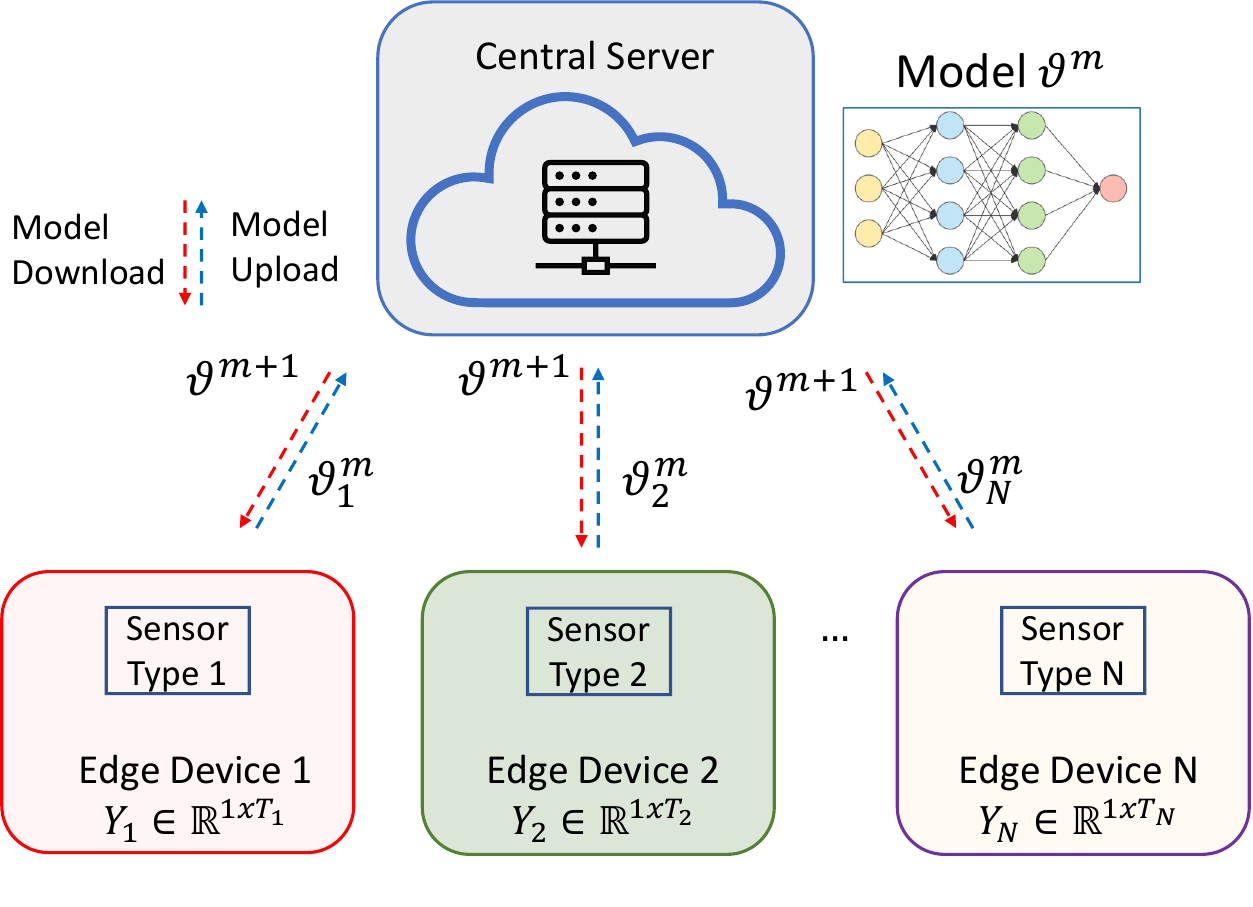}
\caption{The proposed resource-efficient federated learning protocol, where each edge device contains sensors that measure different physical quantities, thus having different feature space. In this illustration, each edge device has access to only one sensor ($\mat{M_i} =\mat{1}$). }
\label{fig:FL_non_iid}
\end{figure*}

\section{Proposed Federated Learning in IoT architecture}
\label{sec:proble_form}
To formulate the examined federated learning framework, a network with  $N$ edge devices is considered. Each edge device $i \in \{1, 2, \dots, N\}$ consists of $M_i$ sensors measuring different physical quantities  e.g., temperature, humidity, energy consumption, e.t.c.. Particularly, each device $i$ has locally a time series dataset $\mat{Y_i} = \{y^1_i, y^2_i, \dots, y_i^{T_i}\}$ comprised of a sequence of $T_i$ measurements. The $t-$th measurement $y^t_i \in \real^K_i$ contain $M_i$ features measured at the time step $t$. 
Figure \ref{fig:FL_non_iid} illustrates the proposed federated architecture, in the case where each edge device consists of only one sensor, thus containing univariate data (i.e., $M_i=1$).

Nonetheless, in IIoT applications, the local time series data $\mat{Y}_i$ of each edge device may contain anomaly  measurements due to faulty sensors and communication failures. 
In centralized solutions, the complete time series dataset from all involved edge devices $\mathcal{\mat{Y}} \in \real^{D\times T} = \{Y_i\}^N_{i=1} $, where $D=N \times M_i$, is required to be gathered in a sever in order to train a machine/deep learning for addressing the anomaly detection and restoration problem. However, this approach
not only introduces a substantial burden on the communication
links between the server and edge devices, as the devices
need to upload their data to server but also may entail risks
regarding the privacy of each device.

To this end, in this study, a novel distributed approach is proposed that pushes all the involved computations toward to the edge. In particular,  taking into consideration, the computational and power limitations of the edge devices \cite{IIoT2, iot}, a resource efficient  federated learning architecture is derived, allowing the edge devices to learn a  compressed global model without sharing their local datasets. Note that in this scenario the edge devices do not share the same feature space (i.e., each edge device may contain different sensors), hence the considered federated learning approach constitutes a non i.i.d. problem.

\section{Proposed resource efficient federated learning methodology}
\label{sec:proposed_model}
In this section, details of the proposed resource efficient federated learning approach is provided, describing the main operations of the involved entities, that is the centralized server and the dispersed edge devices. 

\subsection{Server-side}\label{server}
On the server-side, the server aims to compute a global model by utilizing a fusion rule  that combines all the received local models from the dispersed edge devices. In vanilla FL methodology \cite{fedAvg}, the server employs an average update rule of the edge devices models derived by the following optimization problem.
\begin{equation}
\theta_g^{m}= \arg \min_{\theta_g^{m}} \sum_{i=1}^N \norm{\theta_g^{m} - \theta_i^{m}}^2_\textrm{F}.
\label{eq:aggregation}
\end{equation}
where $\theta_g^{m}$, $\theta_i^{m}$ denote the global model and the local model of the i$-th$ edge device at the m$-th$ communication round. 

However, considering that the limited computational and power resources of the edge devices impose major restrictions during the training and more importantly the inference time, in this study, a model compression fusion rule is proposed, which aims to combine the local models by calculating a  compressed global model and thus  achieving model compression with negligible
accuracy loss. Hence, the proposed optimization problem is described by the following cost function

\begin{equation}
\frac{1}{2} \sum_{i=1}^N \norm{\theta_g^{m} - \theta_i^{m}}^2_\textrm{F} + \lambda \norm{\theta_g^{m}}_1.
\label{eq:compression}
\end{equation}

where $\lambda$ is a positive scalar constants that controls the relative importance of the $l-1$ sparsity imposed regularizer, promoting sparsity in the global model. After solving the proposed
optimization problem, the server conveys the derived compressed global model back to all edge devices, and the next
communication round is performed.

\subsubsection{Efficient ADMM solver }

The proposed compression fusion rule in (\ref{eq:compression}), although convex requires special treatment due to the non-smooth $l-1$ term. In view of this, the ADMM methodology \cite{23Boyd2010} is employed  by introducing an auxiliary variable $Z$ in order to decouple the original problem into two individual sub-problems. Thus, problem (\ref{eq:compression}) can be written as follows,
\begin{align}
&\frac{1}{2} \sum_{i=1}^N \norm{\theta_g^{m} - \theta_i^{m}}^2_\textrm{F} + \lambda \norm{Z}_1 \\
&s.t.\quad  Z = \theta_g^{m} \nonumber
\label{eq:constr_problem}
\end{align}

The corresponding augmented Lagrangian function of problem (\ref{eq:constr_problem}) is given by
\begin{align}
    \mathcal{L} = \frac{1}{2} \sum_{i=1}^N \norm{\theta_g^{m} - \theta_i^{m}}^2_\textrm{F} + \lambda \norm{Z}_1
      &+u^T(Z-\theta_g^{m})\\ \nonumber
      &+ \frac{b}{2}\norm{Z-\theta_g^{m}}
\end{align}
where $u$ denotes the Lagrange multiplier associated with the constraint \cite{23Boyd2010}, and $b>0$ stands for the penalty parameter. 

Hence, a sequence of individual sub-problems emerges, given by,
\begin{align}
    \theta_g^{m,k+1}  &= \underset{\theta_g^{m}}{\arg\min}\,\,\mathcal{L}(\theta_g^{m}, Z^k, u^k) \nonumber\\
    Z^{k+1}  &= \underset{Z}{\arg\min}\,\,\mathcal{L}(\theta_g^{m, k+1}, Z, u^k) \\
    U^{k+1}  &= \underset{U}{\arg\min}\,\,\mathcal{L}(\theta_g^{m, k+1}, Z^{k+1}, u) \nonumber
    \label{eq:subproblems}
\end{align}

The solutions of the above problems are 
\begin{align}
    \theta_g^{m,k+1}  &= \underset{\theta_g^{m}}{\arg\min}\,\,\mathcal{L}(\theta_g^{m}, Z^k, u^k) \nonumber\\
    Z^{k+1}  &= soft\,(\theta_g^{m,k+1} - u^k, \lambda/b ) \\
    u^{k+1}  &= u^k + b(Z^{k+1}-\theta_g^{m,k+1}) \nonumber
    \label{eq:solutions}
\end{align}
where the $soft(.,\tau)$ denotes the soft-thresholding operator $x=sign(x)max(\mid{x}\mid-\tau,0)$. The aforementioned steps are repeated iteratively until convergence is achieved. Once this occurs, the server sends the compressed global model to the edge devices.

\subsection{Edge Devices-side}\label{device}
Focusing on the edge-device side, at every communication round $m$, each  device $i$ receives the compressed global model $\theta_g^m$ and aims to update its local model $\theta_i^m$ by employing its private local time series dataset $\mat{Y}_i$. 

Considering, that the time series data $\mat{Y}_i$ exhibit strong  dependencies  across the dimension of the
time,  we employ the sliding window methodology \cite{window} to effectively capture these dependencies. In more detail, the local time series data $\mat{Y}_i$ of each edge device $i$ is processed into overlapping sequences with time length $w$,  $\{X_i^q\}_{q=1}^Q$. In other words, each derived sequence $X^q_i = \{y_i^t, y_i^{t+1}, \cdots, y_i^{t+w-1} \} \in \real^{M_i \times w}, q=1,\cdots Q$ consists of $w$ measurements of the dataset $\mat{Y}_i$.

Once the local dataset $\mat{X}_i$ is formulated, the edge device proceeds with the training procedure. To ensure that the local  model will remain close to the compressed global model during the training procedure, a regularized  objective function is utilized 
\begin{equation}
  \arg \min_{\theta_i^{m}}  L_i(X_i, \theta^m_i) + \mu \norm{\theta_g^m - \theta^m_i}^2
  \label{eq:loss}
\end{equation}
where $L_i(\cdot)$ denotes a general definition of the loss function describing any supervised/unsupervised learning problem, where its parameters are the local time series dataset and the local model. Additionally, the second term known as proximal regularization term \cite{MLSYS2020_38af8613} is added to the objective function to assist in the compression process performed by the server. Its primary purpose is to keep the local model closely aligned with the compressed global model, hence ensuring that the compression performed at the server remains effective.

It should be highlighted that the above optimization problem is equivalent to the original neural network training plus a $L_2$ regularizer, thus it can be solved employing the stochastic gradient descent, since both terms are differentiable.
After the local updates, the participated devices broadcast their models back to the centralized server.


\subsection{Masked fine-tuning process}

The proposed federated learning framework aims to develop a highly compressed and accurate global model, which may necessitate a significant number of communication rounds. To expedite this process and improve the global model's performance accuracy and convergence, a masked retraining step is implemented. This step involves additional $J$ iterations, known as \textbf{fine-tuning rounds}, between the server and the edge devices.

During the fine-tuning rounds, edge devices are instructed to update only the non-zero weights in their local training process. This selective updating is achieved by applying masks to the gradients of zero weights in the local models, effectively preventing any updates to those weights. Thus, for the rest fine-tuning rounds

On the server-side, the aggregation of local models from edge devices takes into account that only the non-zero weights have been updated during the fine-tuning rounds. Therefore, the server employs a straightforward aggregation fusion rule, as outlined in (\ref{eq:aggregation}), to calculate the new global model while maintaining the zero weights untouched.

The masked retraining communication rounds are performed iteratively until the compressed global model's performance accuracy reaches a satisfactory level. This process ensures that the global model is both resource-efficient and accurate, making it suitable for real-world applications, particularly in resource-constrained edge computing environments.


 
 \textit{Algorithm \ref{algo} summarizes the proposed federated learning approach.}
 
\begin{figure}
\centering
\includegraphics[width=\linewidth]{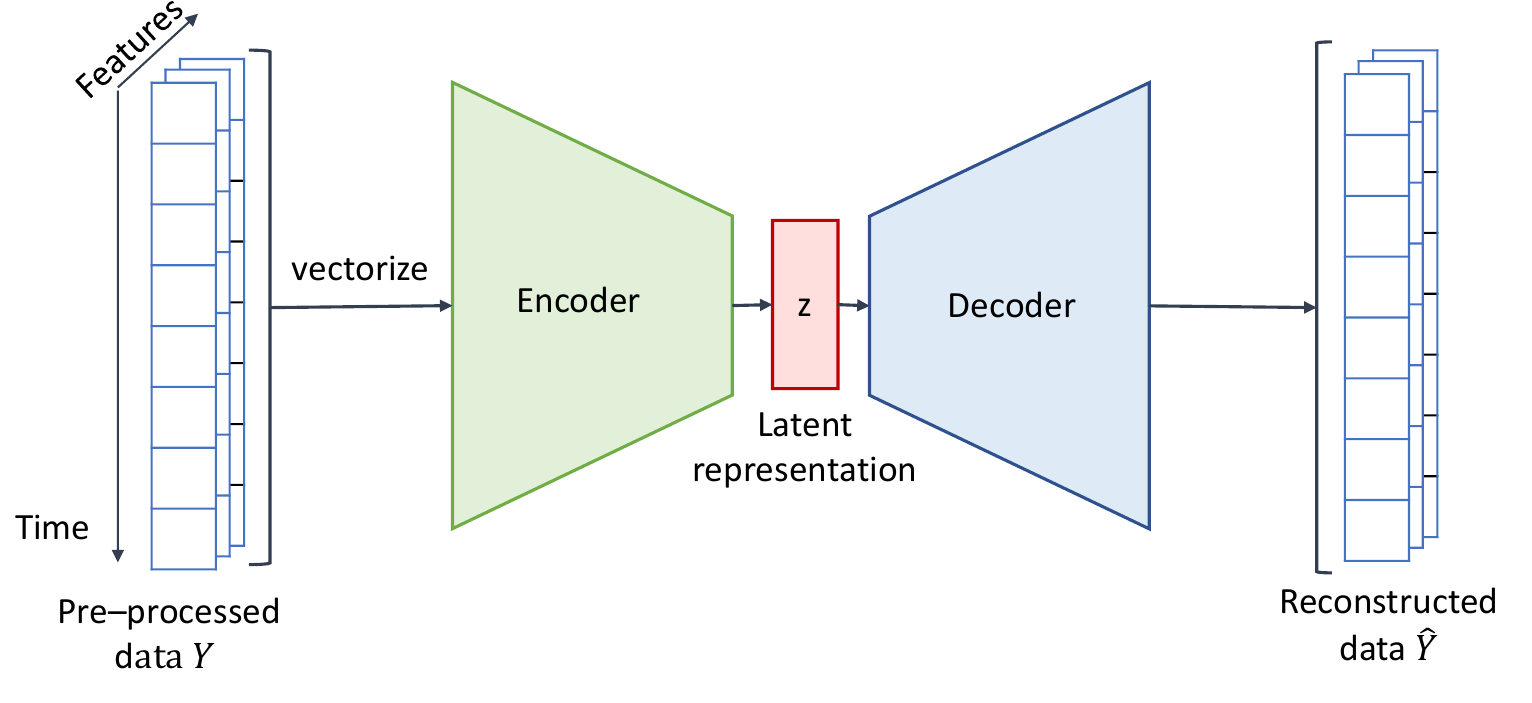}
\caption{The autoencoder-based model employed by the
edge devices utilizing the pre-processed local datasets derived from the time window methodology. }
\label{fig:encoder}
\end{figure}

\subsection{Autoencoder-based model for anomaly detection and restoration}
To address the anomaly detection and restoration problem in multidimensional time series data, an autoencoder-based model is proposed, deployed by the participated edge devices. In general the autoencoder aims to copy its input to its output, by projecting the data into a low dimensional latent space \cite{Ca2010}. Due to their simplicity and low computational complexity, autoencoders are ideal models for edge devices with limited processing capabilities.

To increase the receptive field of the autoencoder, thus capturing the strong time dependencies among the time series data $\mat{Y}_i$, we follow the sliding window approach, detailed in Section \ref{device}.
Having derived the pre-processed local datasets $\mat{X}_i$ , each edge-device $i$ aims to train a local autoencoder-based model utilizing the regularized optimization problem in (\ref{eq:loss}) and employing as loss function in optimization problem (\ref{eq:loss}), the following,
\begin{equation}
    L_i(X_i, \theta^m_i) =  \sum_{p=1}^P\norm{x^q_i - \hat{x}^q_i}_\mathcal{O}^2 = \sum_{q=1}^Q\norm{x^q_i - \mathcal{D}(\mathcal{E}(x^q_i))}_\mathcal{O}^2, 
    \label{eq:modified_auto2}
\end{equation}
where $x^q_i \in \real^{l}, l=M_i \times w$ denotes the vectorized version of the pre-processed local data $X^q_i$, $\mathcal{E(\cdot)}$ denotes the encoder network aiming to to compute
the intrinsic hidden representation of the input data and $\mathcal{D(\cdot)}$ is the decoder network that targets to decode the
derived hidden representation of the encoding process back to the input data. Note that in case where the local datasets contain missing value anomalies the norm $\norm{\cdot}_\mathcal{O}^2$ only considers the contribution of the observed values, ignoring the missing values of the local time series data. Hence, during the training procedure of the local  serialized autoencoder model i.e., $\vartheta^m_i$, the backpropagation updates only those weights associated with the observed values of the input.

\textbf{Detect Anomalies:} Focusing on the anomaly detection task, the above loss function enables the autoencoder-based models of the edge devices to learn the distribution of the normal data inside the local training datasets. Thus,  once the models are trained, they are capable of estimating data points very similar to the training normal data distribution. During the inference stage, the estimated values will follow the distribution of the normal data. Hence, if an anomalous measurement occurs, the trained models will fail to reconstruct it accurately.  In more detail, when the reconstruction error (anomaly score) exceeds a certain threshold $E$, the corresponding data point is determined as an anomaly \cite{anomaly_review}. To define a proper threshold for the anomaly detection approach, we employ the following relation
\begin{equation}
    E = \mu + c \cdot \sigma
\end{equation}
where $\mu$, $\sigma$ denote the mean and variance values of the training data reconstruction error and $c$ is a user-defined parameter that controls the sensitivity of the threshold.

\begin{algorithm}
\caption{: Proposed Compressed-based Federated learning method}
\begin{algorithmic}
\REQUIRE Model, number of communication rounds $M$, fine-tuning rounds J, Loss function based on eq (\ref{eq:loss}).\\

\ENSURE \,\, Global model $\theta^M_g$
\STATE \textbf{------------------- Compression Stage: -------------------}
\STATE \textbf{Server side:}
\STATE Initialize the global model $\theta^0_g$
\FOR {each communication round $m=1:M$}
\STATE Send the global model $\theta^m_g$ to all edge devices
\STATE Wait the uploaded local models $\{\theta_i\}^N_{i=1}$ from all devices
\STATE Compute the new compressed global model $\theta_g^{m+1}$ using the iterative solution of the ADMM solver in (6) derived from the compressed-based optimization problem in (\ref{eq:constr_problem}).
\ENDFOR 
\STATE \textbf{Edge device side:}
\FOR {each communication round $m=1:M$}
\FOR {each device $i=1:N$}
\STATE Initialize the local model $\theta^m_i$ with the   compressed global model $\theta^m_g$
\STATE Train the local model based on optimization problem (\ref{eq:loss})
\STATE Upload the updated local model back to the server
\ENDFOR 
\ENDFOR 

\STATE\textbf{ -------------- Masked fine-tuning stage: ---------------}
\IF {the compression rate of the global model is satisfactory}
\STATE \textbf{Server side:}
\FOR {each fine-tuning round $j=1:J$}
\STATE Send the global model $\theta^j_g$ to all edge devices
\STATE Wait the uploaded local models $\{\theta_i\}^N_{i=1}$ from all devices
\STATE Compute the new global model $\theta_g^{m+1}$ using the aggregation fusion rule in (\ref{eq:aggregation})
\ENDFOR 

\STATE \textbf{Edge device side:}
\FOR {each fine-tuning round $j=1:J$}
\FOR {each device $i=1:N$}
\STATE Initialize the local model $\theta^j_i$ with the  global model $\theta^j_g$
\STATE  The
devices are permitted to update only the non zero weights
during their local training procedure, thus, masks are
applied to gradients of zero weights of the local models
preventing them from updating
\STATE Upload the updated local model back to the server
\ENDFOR 
\ENDFOR 

\ENDIF
\end{algorithmic}
\label{algo}
\end{algorithm}

\begin{figure*}
\centering
\includegraphics[scale=0.4]{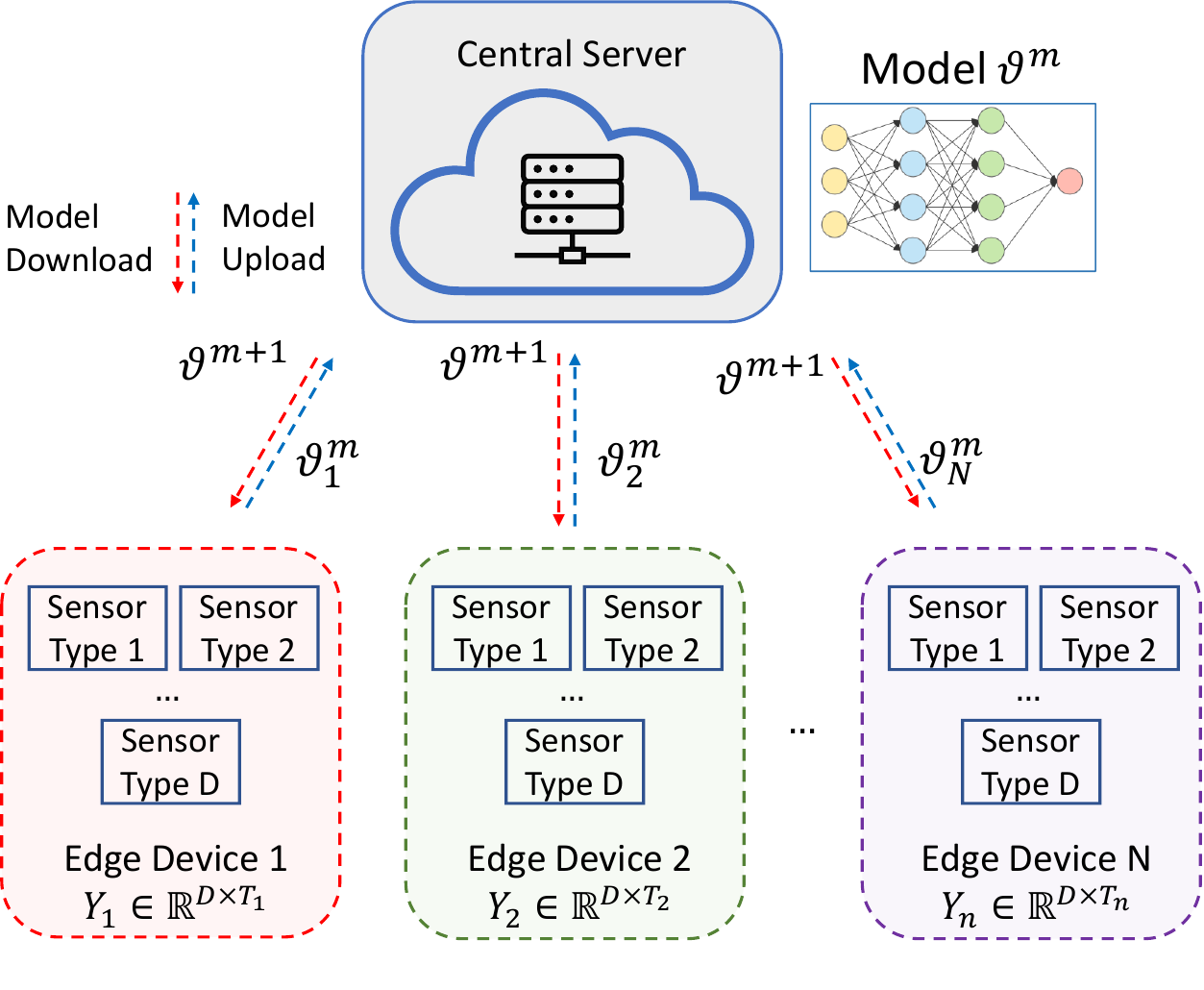}
\caption{The Federated Learning classical scenario (FL-multivariate), where each edge device contains the same type of sensors that measure the exact same quantities.
}
\label{fig:FL_iid}
\end{figure*}

\section{Experimental Part}
To highlight the efficiency and applicability of the proposed resource efficient federated learning framework, extensive experiments were curried out on a real-world multidimensional time series dataset in the context of the missing value imputation and anomaly detection problems.

\textbf{Dataset} \cite{energy_data}:The considered multivariate time series dataset consists of 27 features derived from dispersed wireless sensors measuring various physical quantities e.g., temperature, humidity, pressure, energy consumption from a building. The derived measurements were recorded every $10$ minutes over several months, thus leading to a multidimensional time series dataset $\{\mat{Y}_i\}_{i=1}^N$ with $N=27$ features, where each $\mat{Y}_i \in \real^{1 \times T}$ contains T=19000 measurements. 

\textbf{Compared Methods}: To tackle the problem of anomaly detection and restoration, we considered the following approaches
 \begin{itemize}
  \setlength{\itemsep}{1pt}
  \setlength{\parskip}{0pt}
  \setlength{\parsep}{0pt}
     \item Centralized scenario: The server has access to all local time series datasets of the participated edge devices. Thus, a single model is learnt based on the whole dataset $\mat{Y}$. 
     
     \item Federated learning multivariate scheme (FL-multivariate, see Figure \ref{fig:FL_iid}): To simulate the scenario where the clients have the same feature space (i.e., the same sensors), the whole time series dataset is multivariately distributed to the edge devices. In more detail, we assume that each edge device contains the same types of sensors, thus deriving a local time series dataset with 24 features. Note that the practical value of this scenario is limited, since require from all the involved edge devices to have the same feature space. Thus, the whole time series dataset is split into 5 edge devices, where each device contains 24 features with $(19000/5)$ observations. Note that the above scenario is followed to various studies \cite{fl_a1, fl_a3, fl3, fl4, fl5, fl6, fl7}, as mentioned in Section\ref{related}.
     
     \item Proposed FL approach (FL-univariate, see Figure \ref{fig:FL_non_iid}): In this case, we propose a scenario with more practical value, where the devices have different sensors, and thus measuring different physical quantities. Hence, since the considered dataset contains $24$ features, we employed $N=24$ edge devices with $\mat{M_i = 1}$ sensor. 
     Additionally, utilizing the proposed compressed based optimization problem in (\ref{eq:compression}), the FL-univariate compressed method is derived. The goal is to 
to capture the multivariate information by integrating their
individual univariate data.
 \end{itemize}

\textbf{Parameter Settings}:
Concerning the dataset, a time window with size $w=50$ was employed to split the time series data into overlapping sequences. For the centralized and the FL-multivariate scenario, since the share the same feature space (i.e., the same number of features) an autoencoder with two layers of size $\{128, 64,64, 128\}$ was used. Additionally, for the proposed FL-univariate scenario an autoencoder with size $\{64, 32, 32, 64\}$ was employed.
Regarding the training  of the centralized scenario, we used  $50$ epochs with learning rate equal to 1e-03. For the two FL scenarios, we employed $30$ epochs with learning rate  equal to $1e-03$  during the training of the local models at the edge devices, and $30$ communications rounds.

\subsection{Anomaly Detection and Restoration - Real world dataset}

\textbf{Detect Outliers:}
In this application, the goal is to detect the anomaly measurements in the time series dataset. To this end, we split the dataset into training, validation and test set introducing randomly outlier points per feature (i.e., points that exceed 3 times the maximum value of each feature). Specifically, we explored two anomaly rates  i.e., $\{10\%, 30\% \}$. Table \ref{tab:my_label2} summarizes the anomaly detection results in terms of precision , recall and accuracy metrics. As can be clearly seen the pro-
posed FL-univariate method, although it considers edge-devices
with different sensors, it is able to exhibit competitive performance against the FL-multivariate approach that considers devices
with the same feature space and the centralized solution. Furthermore, the compressed federated learning versions of the
FL-univariate and FL-multivariate approaches are able to achieve high
compression rates without sacrificing accuracy.

\textbf{Missing value Imputation:}
In this IIoT application, the Missing Completely at Random (MCAR) methodology is employed to insert 
missing values in the training, validation and test set. In particular, we examined three missing value rates p, i.e., $\{10\%, 30\%, 50\% \}$. The goal is to learn a model  to predict the missing values in the test set. The performance is estimated based on the ground truth the corresponding imputed values in terms of Root Mean Squared Error (RMSE). 
Table \ref{tab:missing_values} summarizes the imputation performance under several missing rates scenarios. Similar to the previous results, the proposed FL-univariate methodology along with its respective compressed version are able to provide similar results with the centralized approach.

\begin{table*}
    \centering
        \caption{Anomaly detection performance and compression rate of the proposed federated learning approach (i.e., FL-univariate) with N = 24 edge devices (each edge device has only one sensor $\mat{M_i=1}$) compared to the classical federated learning scenario (i.e., FL-multivariate) with N = 5 edge devices  and the centralized scheme.}
    \resizebox{17cm}{!}{
    \begin{tabular}{c c  c c c c c}
    \toprule
    Anomaly &  & Centralized & FL-multivariate & FL-multivariate & \textbf{FL-univariate} & \textbf{FL-univariate}\\
     value rate & & &&compressed&&\textbf{compressed} \\
     \midrule
           &Recall           &0.9985     & 0.9952   & 0.9951     & 0.9947      & 0.991\\
           &Prec             &0.9932     & 0.9901   & 0.9942     & 0.9921     & 0.9903     \\
           &Acc              & 0.9996    & 0.9919   & 0.9905    & 0.9929     & 0.9910     \\
      10\% &No. Para.     &325K&  325K& 19.5K     &  10K    & 0.7K     \\
           &No. features (sensors) per device&-&$M_i=24$& $M_i=24$&$M_i=1$ &$M_i=1$\\
          &Compress. rate &  -    &   -   &   94\%   &  \textbf{96.2\%}    &   \textbf{99.78\%}   \\
     
     \midrule
           &Recall           &0.9924     & 0.9917    & 0.9864     & 0.9832      & 0.9806\\
           &Prec             &0.9883     & 0.9838    & 0.9825    & 0.9803      & 0.9804     \\
           &Acc              & 0.9892    & 0.9845    & 0.9814     & 0.9825      & 0.9800    \\
      30\% &No. Para.        &325K       &325K      & 19.5K      & 10K        & 0.5K     \\
                 &No. features (sensors) per device&-&$M_i=24$& $M_i=24$&$M_i=1$ &$M_i=1$\\
          &Compress. rate    &  -        &   -      &   94\%    & \textbf{96.2\%}         &   \textbf{99.84\%}  \\
     \bottomrule
    \end{tabular}}

    \label{tab:my_label2}
\end{table*}

\begin{table*}
    \centering
        \caption{Imputation performance and compression rate of the proposed federated learning approach (i.e., FL-univariate) with N = 24 edge devices (each edge device has only one sensor $\mat{M_i=1}$) compared to the classical. federated learning scenario with N = 5 edge devices (i.e., FL-multivariate) and the centralized scheme.}
    \resizebox{17cm}{!}{
    \begin{tabular}{c c  c c c c c}
    \toprule
    Missing&  & Centralized & FL-multivariate & FL-multivariate & \textbf{FL-univariate} & \textbf{FL-univariate}\\
     value rate & &  &.&compressd&&\textbf{compressed} \\
     \midrule
           &RMSE          &10.23 & 10.58& 10.69& 10.74& 10.81\\
      10\% &No. Para.     &325K&  325K& 26K     &  10K    & 0.7K     \\
                 &No. features (sensors) per device&-&$M_i=24$& $M_i=24$&$M_i=1$ &$M_i=1$\\
           &Compress. rate &  -    &   -   &   \textbf{92\%}   &  \textbf{96.2\%}   &   \textbf{99.78\%}   \\
     
     \midrule
           &RMSE         &12.21&12.53&12.62 & 12.64& 12.69\\
      30\% &No. Para.     &325K&  325K& 26K     &  10K    & 0.6K     \\
                 &No. features (sensors) per device&-&$M_i=24$& $M_i=24$&$M_i=1$ &$M_i=1$\\
           &Compress. rate &  -    &   -   &   \textbf{92\%}   & \textbf{96.2\%}    &   \textbf{99.81\%}     \\
     \midrule
           &RMSE         &13.16 & 13.62& 13.69& 13.81& 13.88\\
      50\% &No. Para.    &325K&  325K& 33K     &  10K    & 0.8K    \\
                 &No. features (sensors) per device&-&$M_i=24$& $M_i=24$&$M_i=1$ &$M_i=1$\\
           &Compress. rate &  -    &   -   &   \textbf{90\%}   &  \textbf{96.2\%}    &   \textbf{99.75\%}  \\
     \bottomrule
    \end{tabular}}

    \label{tab:missing_values}
\end{table*} 

\begin{figure}
\centering
\includegraphics[scale=0.5]{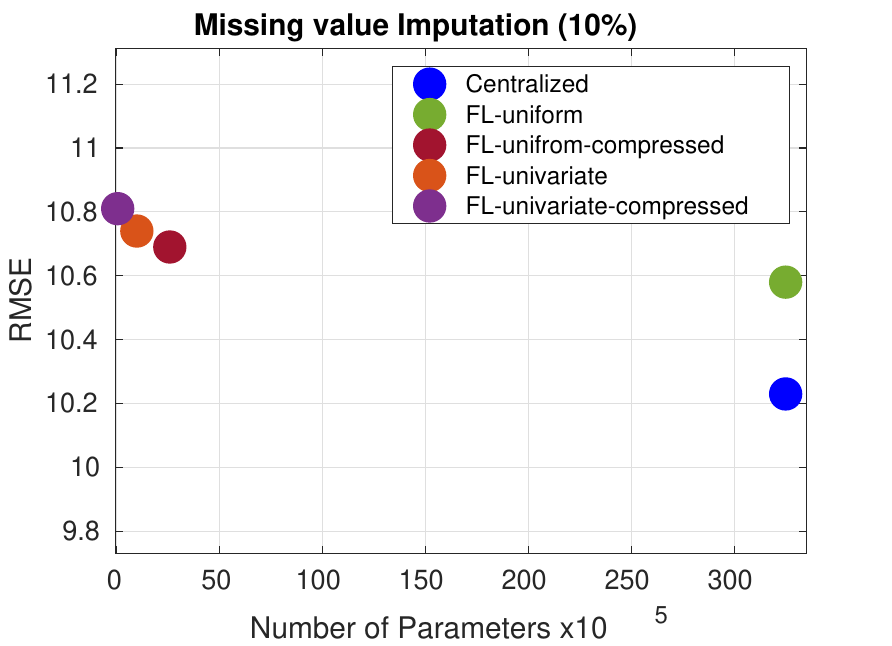}
     \caption{Model complexities comparison of our proposed federated learning scenario (FL-univariate) with and without the compression scheme, the classical FL scenario with and without the compression scheme and the centralized solution. 
}
  \label{fig:num_of_param}
\end{figure}

Overall, considering Table  \ref{tab:my_label2} and \ref{tab:missing_values}, it can be deduced that utilizing  the proposed FL-multivariate scenario, we can reduce the number of parameters of the global model by $\textbf{96.2\%}$ compared to models derived by the centralized and FL-multivariate solutions, since the local
datasets  of the edge devices have smaller feature space (one feature/sensor) compared to the other two approaches. Additionally, by combining the FL-univariate method with the proposed compressed-based strategy (termed as FL-univariate compressed method), a compression rate greater than $\mat{99,7\%}$ can be achieved  without sacrificing any performance accuracy, as we can see in Figure \ref{fig:num_of_param}. In more detail, in the anomaly detection task, the performance degradation of the FL-univariate -compressed scheme is negligible (less than $1.18\%$), whereas in the missing imputation task the performance loss is less than $5\%$ as compared with the performances of the centralized solution.  Another great benefit that stems out from the proposed FL-univariate and FL-univariate compressed schemes is the fact that these approaches are  able to provide competitive results even if the edge devices train their local models using only a subset of the multidimensional time series data (they have access to univariate data derived from only one sensor), contrary to the centralized and  FL-multivariate  solutions that have access to the whole feature space of the data. 
Thus, the proposed FL-univariate compressed method can be considered ideal  for IIoT edge devices with limited computational and power resources.

\section{Conclusions}
\label{sec:conclu}
In this work, the problem of anomaly detection and restoration was studied under a resource efficient federated learning perspective. To overcome the limitations of the centralized solutions, the the
proposed federated learning scheme pushes all the involved computations at the edge, where each edge device measures only  a part of a multidimensional time series. In addition, considering the limited computational and power resources of the IIoT edge devices, a novel compression-based optimization problem is proposed at the server-side in order to fuse the local models broadcast by the edge devices, thus deriving a compressed global model with reduced number of weights without simultaneously affecting its performance accuracy.
Extensive
experiments were performed on a real-world time series dataset, examining the missing value imputation and anomaly detection problems
 to highlight the efficiency
and applicability of the proposed compressed-based federated learning framework. The proposed framework is able to achieve \textbf{compression rates greater than $\mat{99.7}\%$} without any degradation to its performance.

\bibliographystyle{IEEEtran}
\bibliography{mybibfile}

\begin{thebibliography}{10}
\providecommand{\url}[1]{#1}
\csname url@samestyle\endcsname
\providecommand{\newblock}{\relax}
\providecommand{\bibinfo}[2]{#2}
\providecommand{\BIBentrySTDinterwordspacing}{\spaceskip=0pt\relax}
\providecommand{\BIBentryALTinterwordstretchfactor}{4}
\providecommand{\BIBentryALTinterwordspacing}{\spaceskip=\fontdimen2\font plus
\BIBentryALTinterwordstretchfactor\fontdimen3\font minus
  \fontdimen4\font\relax}
\providecommand{\BIBforeignlanguage}[2]{{%
\expandafter\ifx\csname l@#1\endcsname\relax
\typeout{** WARNING: IEEEtran.bst: No hyphenation pattern has been}%
\typeout{** loaded for the language `#1'. Using the pattern for}%
\typeout{** the default language instead.}%
\else
\language=\csname l@#1\endcsname
\fi
#2}}
\providecommand{\BIBdecl}{\relax}
\BIBdecl

\bibitem{IIoT1}
L.~D. Xu, W.~He, and S.~Li, ``Internet of things in industries: A survey,''
  \emph{IEEE Transactions on Industrial Informatics}, vol.~10, no.~4, pp.
  2233--2243, 2014.

\bibitem{IIoT2}
E.~Sisinni, A.~Saifullah, S.~Han, U.~Jennehag, and M.~Gidlund, ``Industrial
  internet of things: Challenges, opportunities, and directions,'' \emph{IEEE
  Transactions on Industrial Informatics}, vol.~14, no.~11, pp. 4724--4734,
  2018.

\bibitem{anomaly_review}
K.~Choi, J.~Yi, C.~Park, and S.~Yoon, ``Deep learning for anomaly detection in
  time-series data: Review, analysis, and guidelines,'' \emph{IEEE Access},
  vol.~9, pp. 120\,043--120\,065, 2021.

\bibitem{anomaly}
A.~Deng and B.~Hooi, ``{Graph Neural Network-Based Anomaly Detection in
  Multivariate Time Series},'' \emph{Proceedings of the AAAI Conference on
  Artificial Intelligence}, vol.~35, no.~5, pp. 4027--4035, may 2021.

\bibitem{multidimensional}
P.~Bansal, P.~Deshpande, and S.~Sarawagi, ``Missing value imputation on
  multidimensional time series,'' \emph{Proc. VLDB Endow.}, vol.~14, no.~11, p.
  2533–2545, jul 2021.

\bibitem{9976093}
A.~Gkillas and A.~S. Lalos, ``Missing data imputation for multivariate time
  series in industrial iot: A federated learning approach,'' in \emph{2022 IEEE
  20th International Conference on Industrial Informatics (INDIN)}, 2022, pp.
  87--94.

\bibitem{classification}
A.~Gupta, H.~P. Gupta, B.~Biswas, and T.~Dutta, ``An unseen fault
  classification approach for smart appliances using ongoing multivariate time
  series,'' \emph{IEEE Transactions on Industrial Informatics}, vol.~17, no.~6,
  pp. 3731--3738, 2021.

\bibitem{forecasting}
N.~Jin, Y.~Zeng, K.~Yan, and Z.~Ji, ``Multivariate air quality forecasting with
  nested long short term memory neural network,'' \emph{IEEE Transactions on
  Industrial Informatics}, vol.~17, no.~12, pp. 8514--8522, 2021.

\bibitem{nature}
Z.~Che, S.~Purushotham, K.~Cho, D.~Sontag, and Y.~Liu, ``{Recurrent Neural
  Networks for Multivariate Time Series with Missing Values},''
  \emph{Scientific Reports}, vol.~8, no.~1, pp. 1--12, dec 2018.

\bibitem{GANa}
D.~Li, D.~Chen, B.~Jin, L.~Shi, J.~Goh, and S.~K. Ng, ``{MAD-GAN: Multivariate
  Anomaly Detection for Time Series Data with Generative Adversarial
  Networks},'' \emph{Lecture Notes in Computer Science (including subseries
  Lecture Notes in Artificial Intelligence and Lecture Notes in
  Bioinformatics)}, vol. 11730 LNCS, pp. 703--716, 2019.

\bibitem{transformer}
P.~Bansal, P.~Deshpande, and S.~Sarawagi, ``Missing value imputation on
  multidimensional time series,'' \emph{Proc. VLDB Endow.}, vol.~14, no.~11, p.
  2533–2545, jul 2021.

\bibitem{cnn_anomaly}
\BIBentryALTinterwordspacing
T.~Wen and R.~Keyes, ``Time series anomaly detection using convolutional neural
  networks and transfer learning,'' 2019. [Online]. Available:
  \url{https://arxiv.org/abs/1905.13628}
\BIBentrySTDinterwordspacing

\bibitem{AnomalyInd}
W.~Zhang, X.~Li, H.~Ma, Z.~Luo, and X.~Li, ``{Federated learning for machinery
  fault diagnosis with dynamic validation and self-supervision},''
  \emph{Knowledge-Based Systems}, vol. 213, p. 106679, feb 2021.

\bibitem{AnomalyIoT}
Y.~Liu, S.~Garg, J.~Nie, Y.~Zhang, Z.~Xiong, J.~Kang, and M.~S. Hossain, ``Deep
  anomaly detection for time-series data in industrial iot: A
  communication-efficient on-device federated learning approach,'' \emph{IEEE
  Internet of Things Journal}, vol.~8, no.~8, pp. 6348--6358, 2021.

\bibitem{McMahan2017CommunicationEfficientLO}
H.~B. McMahan, E.~Moore, D.~Ramage, S.~Hampson, and B.~A. y~Arcas,
  ``Communication-efficient learning of deep networks from decentralized
  data,'' in \emph{AISTATS}, 2017.

\bibitem{icassp1}
J.~Ding, E.~Tramel, A.~K. Sahu, S.~Wu, S.~Avestimehr, and T.~Zhang, ``Federated
  learning challenges and opportunities: An outlook,'' in \emph{ICASSP 2022 -
  2022 IEEE International Conference on Acoustics, Speech and Signal Processing
  (ICASSP)}, 2022, pp. 8752--8756.

\bibitem{10314010}
A.~Gkillas, A.~S. Lalos, E.~K. Markakis, and I.~Politis, ``A federated deep
  unrolling method for lidar super-resolution: Benefits in slam,'' \emph{IEEE
  Transactions on Intelligent Vehicles}, vol.~9, no.~1, pp. 199--215, 2024.

\bibitem{10167897}
C.~Anagnostopoulos, A.~Gkillas, N.~Piperigkos, and A.~S. Lalos, ``Federated
  deep feature extraction-based slam for autonomous vehicles,'' in \emph{2023
  24th International Conference on Digital Signal Processing (DSP)}, 2023, pp.
  1--5.

\bibitem{fl_a1}
W.~Zhang, X.~Li, H.~Ma, Z.~Luo, and X.~Li, ``{Federated learning for machinery
  fault diagnosis with dynamic validation and self-supervision},''
  \emph{Knowledge-Based Systems}, vol. 213, p. 106679, feb 2021.

\bibitem{fl_a3}
X.~Zhou, X.~Liu, G.~Lan, and J.~Wu, ``{Federated conditional generative
  adversarial nets imputation method for air quality missing data},''
  \emph{Knowledge-Based Systems}, vol. 228, p. 107261, 2021.

\bibitem{fl3}
\BIBentryALTinterwordspacing
R.~A. Sater and A.~B. Hamza, ``A federated learning approach to anomaly
  detection in smart buildings,'' \emph{ACM Trans. Internet Things}, vol.~2,
  no.~4, aug 2021. [Online]. Available: \url{https://doi.org/10.1145/3467981}
\BIBentrySTDinterwordspacing

\bibitem{fl4}
\BIBentryALTinterwordspacing
H.~T. Truong, B.~P. Ta, Q.~A. Le, D.~M. Nguyen, C.~T. Le, H.~X. Nguyen, H.~T.
  Do, H.~T. Nguyen, and K.~P. Tran, ``{Light-weight federated learning-based
  anomaly detection for time-series data in industrial control systems},''
  \emph{Computers in Industry}, vol. 140, p. 103692, 2022. [Online]. Available:
  \url{https://www.sciencedirect.com/science/article/pii/S0166361522000896}
\BIBentrySTDinterwordspacing

\bibitem{fl5}
D.~H. Tran, V.~L. Nguyen, I.~B. K.~Y. Utama, and Y.~M. Jang, ``An improved
  sensor anomaly detection method in iot system using federated learning,'' in
  \emph{2022 Thirteenth International Conference on Ubiquitous and Future
  Networks (ICUFN)}, 2022, pp. 466--469.

\bibitem{fl6}
W.~Zhu, D.~Song, Y.~Chen, W.~Cheng, B.~Zong, T.~Mizoguchi, C.~Lumezanu,
  H.~Chen, and J.~Luo, ``Deep federated anomaly detection for multivariate time
  series data,'' in \emph{2022 IEEE International Conference on Big Data (Big
  Data)}, 2022, pp. 1--10.

\bibitem{fl7}
Y.~Liu, S.~Garg, J.~Nie, Y.~Zhang, Z.~Xiong, J.~Kang, and M.~S. Hossain,
  ``{Deep Anomaly Detection for Time-Series Data in Industrial IoT: A
  Communication-Efficient On-Device Federated Learning Approach},'' \emph{IEEE
  Internet of Things Journal}, vol.~8, no.~8, pp. 6348--6358, apr 2021.

\bibitem{8253600}
Y.~Cheng, D.~Wang, P.~Zhou, and T.~Zhang, ``Model compression and acceleration
  for deep neural networks: The principles, progress, and challenges,''
  \emph{IEEE Signal Processing Magazine}, vol.~35, no.~1, pp. 126--136, 2018.

\bibitem{Ac1}
S.~Nousias, E.~V. Pikoulis, C.~Mavrokefalidis, and A.~S. Lalos, ``Accelerating
  deep neural networks for efficient scene understanding in automotive
  cyber-physical systems,'' in \emph{2021 4th IEEE International Conference on
  Industrial Cyber-Physical Systems (ICPS)}, 2021, pp. 63--69.

\bibitem{Ac2}
T.~Zhang, S.~Ye, K.~Zhang, J.~Tang, W.~Wen, M.~Fardad, and Y.~Wang, ``A
  systematic dnn weight pruning framework using alternating direction method of
  multipliers,'' in \emph{Proceedings of the European Conference on Computer
  Vision (ECCV)}, 2018, pp. 184--199.

\bibitem{Ac3}
B.~Liu, M.~Wang, H.~Foroosh, M.~Tappen, and M.~Pensky, ``Sparse convolutional
  neural networks,'' in \emph{Proceedings of the IEEE conference on computer
  vision and pattern recognition}, 2015, pp. 806--814.

\bibitem{10290057}
A.~Gkillas, D.~Ampeliotis, and K.~Berberidis, ``Deep equilibrium models meet
  federated learning,'' in \emph{2023 31st European Signal Processing
  Conference (EUSIPCO)}, 2023, pp. 1873--1877.

\bibitem{iot}
J.~Tang, D.~Sun, S.~Liu, and J.-L. Gaudiot, ``Enabling deep learning on iot
  devices,'' \emph{Computer}, vol.~50, no.~10, pp. 92--96, 2017.

\bibitem{fedAvg}
B.~McMahan, E.~Moore, D.~Ramage, S.~Hampson, and B.~A.~y. Arcas,
  ``{Communication-Efficient Learning of Deep Networks from Decentralized
  Data},'' in \emph{Proceedings of the 20th International Conference on
  Artificial Intelligence and Statistics}, ser. Proceedings of Machine Learning
  Research, A.~Singh and J.~Zhu, Eds., vol.~54.\hskip 1em plus 0.5em minus
  0.4em\relax PMLR, 20--22 Apr 2017, pp. 1273--1282.

\bibitem{eldar}
T.~Gafni, N.~Shlezinger, K.~Cohen, Y.~C. Eldar, and H.~V. Poor, ``Federated
  learning: A signal processing perspective,'' 2021.

\bibitem{23Boyd2010}
S.~Boyd, N.~Parikh, E.~Chu, J.~Eckstein, S.~Boyd, N.~Parikh, E.~Chu,
  B.~Peleato, and J.~Eckstein, ``{Distributed Optimization and Statistical
  Learning via the Alternating Direction Method of Multipliers},''
  \emph{Foundations and Trends R in Machine Learning}, vol.~3, no.~1, pp.
  1--122, 2010.

\bibitem{window}
H.~S. Hota, R.~Handa, and A.~K. Shrivas, ``Time series data prediction using
  sliding window based rbf neural network,'' 2017.

\bibitem{MLSYS2020_38af8613}
T.~Li, A.~K. Sahu, M.~Zaheer, M.~Sanjabi, A.~Talwalkar, and V.~Smith,
  ``Federated optimization in heterogeneous networks,'' in \emph{Proceedings of
  Machine Learning and Systems}, I.~Dhillon, D.~Papailiopoulos, and V.~Sze,
  Eds., vol.~2, 2020, pp. 429--450.

\bibitem{Ca2010}
P.~V. Ca, L.~T. Edu, I.~Lajoie, Y.~B. Ca, and P.-A.~M. Ca, ``{Stacked Denoising
  Autoencoders: Learning Useful Representations in a Deep Network with a Local
  Denoising Criterion Pascal Vincent Hugo Larochelle Yoshua Bengio
  Pierre-Antoine Manzagol},'' Tech. Rep., 2010.

\bibitem{energy_data}
L.~M. Candanedo, V.~Feldheim, and D.~Deramaix, ``{Data driven prediction models
  of energy use of appliances in a low-energy house},'' \emph{Energy and
  Buildings}, vol. 140, pp. 81--97, apr 2017.

\end{thebibliography}

\EOD

\end{document}